\documentclass[10pt,twocolumn,letterpaper]{article}

\usepackage[pagenumbers]{cvpr} 

\usepackage{bbding}
\usepackage{booktabs}
\usepackage{amsmath}
\usepackage{amssymb}
\usepackage{makecell}
\usepackage{multirow}
\usepackage[table]{xcolor}
\usepackage{graphicx}
\usepackage{algpseudocode}
\usepackage[misc]{ifsym} 

\makeatletter
\newcommand\blfootnote[1]{%
  \begingroup
  \def\@makefntext##1{%
    \noindent ##1
  }%
  \footnotetext{#1}%
  \endgroup
}
\makeatother

\usepackage{bm} 
\usepackage[ruled,vlined, linesnumbered, noend]{algorithm2e} 

\usepackage[accsupp]{axessibility}  

\definecolor{cvprblue}{rgb}{0.21,0.49,0.74}
\usepackage[pagebackref,breaklinks,colorlinks,allcolors=cvprblue]{hyperref}

\def\paperID{40039} 
\def\confName{CVPR}
\def\confYear{2026}

\title{Visual Prototype Conditioned Focal Region Generation\\for UAV-Based Object Detection}

\author{
Wenhao Li\textsuperscript{1,2},
Zimeng Wu\textsuperscript{1,2},
Yu Wu\textsuperscript{1,2},
Zehua Fu\textsuperscript{3},
Jiaxin Chen\textsuperscript{1,2 \Letter}\\
\textsuperscript{1}State Key Laboratory of Virtual Reality Technology and Systems, Beihang University, China\\
\textsuperscript{2}School of Computer Science and Engineering, Beihang University, Beijing, China\\
\textsuperscript{3}Hangzhou Innovation Institute, Beihang University, Hangzhou, China\\
{\tt\small \{liwenha0, zimengwu, wuyu22230605, zehua\_fu, jiaxinchen\}@buaa.edu.cn}
}

\begin{document}
\maketitle
\blfootnote{\textsuperscript{\Letter} Corresponding Author}
\begin{abstract}
Unmanned aerial vehicle (UAV) based object detection is a critical but challenging task, when applied in dynamically changing scenarios with limited annotated training data. Layout-to-image generation approaches have proved effective in promoting detection accuracy by synthesizing labeled images based on diffusion models. However, they suffer from frequently producing artifacts, especially near layout boundaries of tiny objects, thus substantially limiting their performance. To address these issues, we propose UAVGen, a novel layout-to-image generation framework tailored for UAV-based object detection. Specifically, UAVGen designs a Visual Prototype Conditioned Diffusion Model (VPC-DM) that constructs representative instances for each class and integrates them into latent embeddings for high-fidelity object generation. Moreover, a Focal Region Enhanced Data Pipeline (FRE-DP) is introduced to emphasize object-concentrated foreground regions in synthesis, combined with a label refinement to correct missing, extra and misaligned generations. Extensive experimental results demonstrate that our method significantly outperforms state-of-the-art approaches, and consistently promotes accuracy when integrated with distinct detectors. 
The source code is available at {\small \url{https://github.com/Sirius-Li/UAVGen}}.
\end{abstract}
\section{Introduction}
\label{sec:intro}

Unmanned Aerial Vehicle (UAV) based object detection~\cite{redmon2016you, li2020generalized, huang2022ufpmp, du2023adaptive, wu2024domain, li2025remdet}, which localizes and identifies targets from aerial perspectives, has demonstrated practical value across a range of applications~\cite{erdelj2017help,honkavaara2013processing, rovira2008stereo,huang2021decentralized}. 
However, as deep learning has become the mainstream paradigm, UAV-based detection remains fundamentally constrained by the scarcity of high-quality training data, especially in dynamically changing environments~\cite{du2019visdrone, du2018unmanned, dieter2023quantifying}. Consequently, its robustness and generalization capabilities still fall short of those achieved in general object detection tasks.

\begin{figure}[t]
\centering
\includegraphics[width=1.0 \columnwidth]{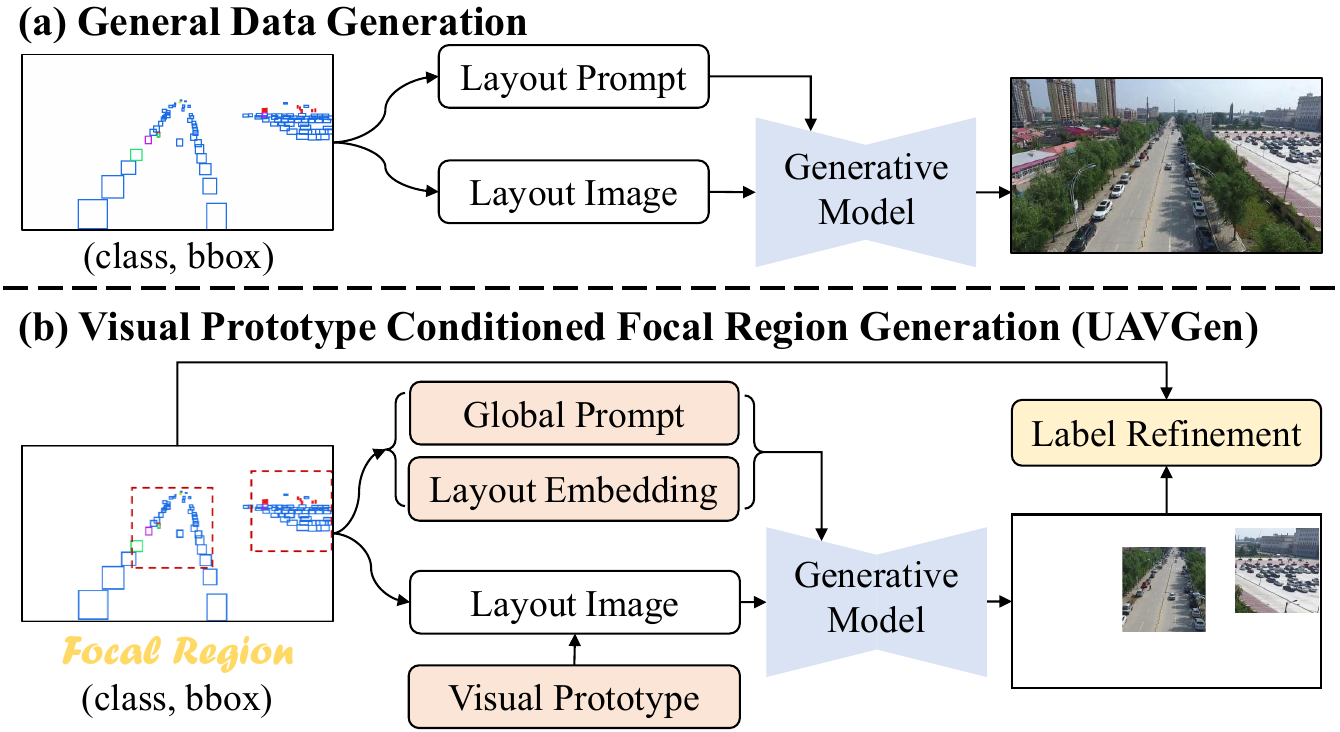} 
\caption{Illustration of different layout-to-images data generation methods. (a) In general data generation, the layout information is directly derived from real data, and synthesized data is directly used as training data for the detector. (b) Our UAVGen enhances this paradigm by introducing multi-cue, high-quality layout conditions for focal regions, as well as refining the synthesized data.}
\label{fig1}
\end{figure}

In order to obtain sufficient and diverse training data under limited cost, data augmentation has emerged as a promising solution~\cite{dwibedi2017cut, gao2023magicdrive, wu2023datasetdm}. Early approaches construct additional samples by applying geometric transformations to existing images, such as flipping, rotation or simple cropping and paste~\citep{ghiasi2021simple, zhao2023x}. 
Recently, benefiting from the generalization capabilities of generative models, diffusion-driven data augmentation has emerged as a new paradigm that synthesizes images from input layouts (referred to as layout-to-image), producing entirely novel variations in appearance. By textual prompt conditioning~\citep{chen2024geodiffusion,wang2024detdiffusion}, latent-space injection~\citep{tang2025aerogen}, or layout-image conditioning~\citep{zhu2024odgen}, it has prone effective in enhancing detector training under limited-data scenarios. Nonetheless, despite their success on general object detection benchmarks, such approaches have not yet achieved comparable gains in UAV-based detection.

We attribute this performance gap to three key limitations inherent to UAV image characteristics:
1) \textit{Low-quality visual layouts with small and overlapped targets.} 
Limited by flight altitude and fixed viewing geometry, UAV-captured objects often present small scales and frequent overlaps. Such regions tend to produce blurred or visually entangled layout representations, which impair the clarity of fine-grained conditional signals, interfere with diffusion model training, and lower the fidelity of synthesized samples.
2) \textit{Inefficient utilization of model capacity under highly uneven spatial distribution.} 
Objects in UAV imagery are typically concentrated in a small fraction of the scene, leaving large regions with negligible informative. Diffusion models inadvertently allocate excessive capacity to these low-information regions, weakening their ability to capture fine-grained object characteristics.
3) \textit{Inconsistency between synthesized images and ground-truth annotations.} Due to the inherent stochasticity of the diffusion process, existing methods inevitably generate images deviate from the input layouts, resulting in missing generations, extra generations and label misalignments. Such inconsistencies are further amplified in small-object-dominated UAV scenarios, introducing label noise that impairs the training of detectors.

To address these challenges, we propose UAVGen, a novel diffusion-driven data augmentation framework for UAV-based object detection. To improve layout condition quality, we first design a visual prototype conditioned diffusion model (VPC-DM). Specifically, a dual-criteria selection mechanism extracts visually clear, semantically well-defined objects from annotated data to construct high-fidelity visual prototypes. With these prototypes, we build layout images integrated with global and fine-grained textual layout semantics, delivering more robust guidance for image generation. 
Second, we develop a focal region enhanced data pipeline (FRE-DP). Instead of conventional full-image generation, it focuses on densely clustered small-object regions, steering both diffusion model and detector optimization toward detection-critical areas. Finally, a label refinement step corrects annotations of synthesized images, providing more precise training data for detectors.

Our main contributions are summarized in three-fold:
\begin{itemize}
    \item We propose a novel framework to synthesize data for UAV-based object detection, dubbed UAVGen. To the best of our knowledge, this work is the first data synthesis approach for UAV-based detector training. 
    \item We design a visual prototype conditioned diffusion model as well as a focal region enhanced data pipeline, to address the problems of low layout-quality, misfocused optimization and image-label inconsistency, thereby facilitating the UAV-based detector training.
    \item Comprehensive experiments on multiple benchmarks demonstrate that our method consistently improves object detection performance, achieving state-of-the-art results across both general and UAV-based detectors.
\end{itemize}
\section{Related Work}
\label{sec:relatedwork}

\begin{figure*}[t]
\centering
\includegraphics[width=0.95\textwidth]{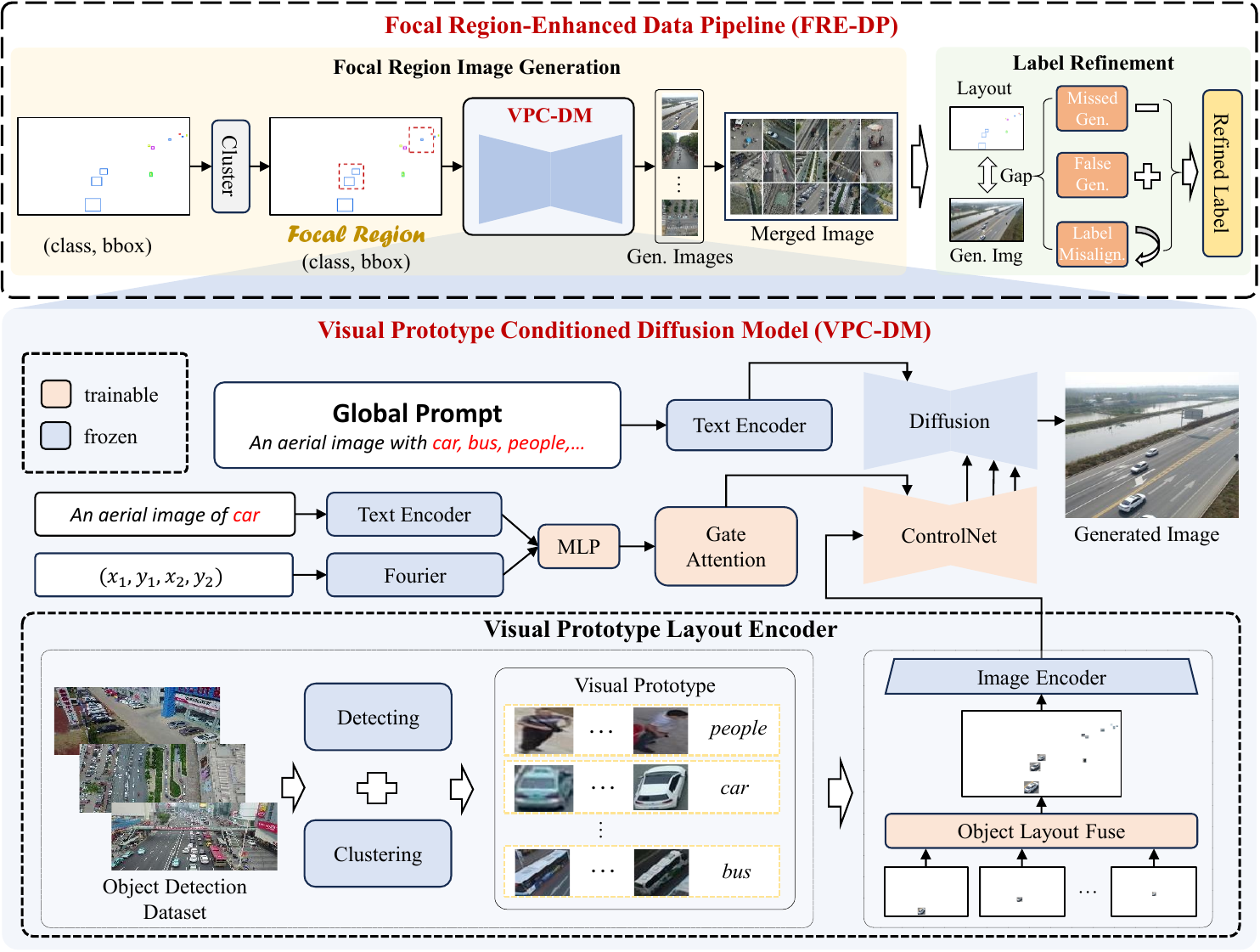}
\caption{\textbf{Architecture of Visual Prototype Conditioned Focal Region Generation.} (a) Virtual Prototype Conditioned Diffusion Model (VPC-DM) generates images guided by layout images which is produced from selected visual prototypes. (b) Focal Region-Enhanced Data Pipeline (FRE-DP) synthesizes images on object-centric areas to avoid limitation of small object generation. Moreover, Label Refinement mitigates the misalignment between layouts and generated images.}
\label{FigMethod}
\end{figure*}

\subsection{UAV-based Object Detection}
Object detection in UAV aerial imagery faces severe challenges including tiny objects, complex backgrounds, and strict computational constraints. Although general-purpose detectors such as the YOLO series~\cite{redmon2016you, redmon2018yolov3, khanam2024yolov11, tian2025yolov12}
and GFL~\cite{li2020generalized} provide efficient foundational frameworks, they often underperform in aerial contexts without domain-specific modifications. To mitigate these issues, various specialized detectors have been developed. CEASC~\cite{du2023adaptive} employs global context-enhanced normalization and adaptive sparse convolution for efficient small object detection. UFPMP-Det \cite{huang2022ufpmp} utilizes foreground packing and multi-proxy networks to jointly optimize accuracy and speed. RemDet \cite{li2025remdet} adopts residual context mining  to enhance the representation of tiny-objects features and reduce the decay of deep-layer information.

\subsection{Diffusion Models} 
Diffusion models have emerged as a leading paradigm in generative modeling due to their stability and superior image synthesis capabilities \citep{ho2020denoising, dhariwal2021diffusion, kingma2021variational, nichol2021improved, rombach2022high}. Unlike earlier frameworks such as Generative Adversarial Networks (GANs) \citep{goodfellow2014generative, brock2018large, karras2019style} and Variational Autoencoders (VAEs) \citep{kingma2013auto, rezende2014stochastic, vahdat2020nvae}, diffusion models generate data through a progressive denoising process. They iteratively refine samples from Gaussian noise to approximate the data distribution \citep{sohl2015deep, song2020denoising, song2020improved}. Recent efforts focus on improving both efficiency and controllability. Latent Diffusion Models (LDMs) \citep{rombach2022high} perform denoising in latent space, boosting computational efficiency while preserving image fidelity. Extensions like ControlNet \citep{zhang2023adding} enable structural conditioning using segmentation masks and layout annotations, allowing finer control over outputs.

\subsection{Layout-to-Image Generation} 
Generating photorealistic images from abstract layouts has attracted increasing attention. Early layout-to-image (L2I) methods based on GANs~\cite{li2021image} or transformers~\citep{jahn2021high}, conditioned on bounding boxes and labels, often suffered from instability and limited realism in complex scenes \citep{li2021image}. Diffusion-based approaches alleviate these issues by enhancing both fidelity and spatial alignment. LayoutDiffusion \citep{zheng2023layoutdiffusion} and LayoutDiffuse \citep{cheng2023layoutdiffuse} incorporate layout-aware attention into LDMs to better model object interactions. Other frameworks, including MultiDiffusion \citep{bar2023multidiffusion}, BoxDiff \citep{xie2023boxdiff}, GLIGEN \citep{li2023gligen}, ReCo \citep{yang2023reco}, and GeoDiffusion \citep{chen2024geodiffusion}, further improve layout conditioning via training-free sampling or architectural enhancements. On a finer level, MIGC \citep{zhou2024migc} and InstanceDiffusion \citep{wang2024instancediffusion} introduce instance-aware control using boxes, masks, or scribbles, enabling more diverse and spatially coherent synthesis.

\subsection{Synthetic Data Generation} 
Copy-Paste methods~\citep{dvornik2018modeling, dwibedi2017cut, ghiasi2021simple, zhao2023x} synthesize training data by randomly placing foreground objects onto background images, but produce noticeable artifacts along object boundaries. Rendering-based methods~\cite{khose2024skyscenes, yim2024synplay, rizzoli2023syndrone, shen2023progressive} synthesize images from 3D simulators to enhance detection, which suffer from domain gap between rendered and real UAV imagery. Recent efforts turn to diffusion-based models to mitigate these limitations. DatasetGAN~\citep{zhang2021datasetgan}, DatasetDM~\citep{wu2023datasetdm}, and Satsynth~\citep{toker2024satsynth} produce high-quality annotated data for object detection. However, challenges remain for small objects, image blurriness, and dense backgrounds—particularly in UAV imagery. Task-specific methods like TrackDiffusion~\citep{li2025trackdiffusion}, MagicDrive~\citep{gao2023magicdrive}, and AeroGen~\citep{tang2025aerogen} improve spatial consistency through multi-object tracking and 3D perception. Yet, occlusion, interference, and scale variation still pose obstacles in aerial scenes.

\section{Method}
\label{sec:method}

\subsection{Framework Overview}
\label{sec:framework_overview}
We follow the typical paradigm of diffusion-driven data augmentation for object detection, conducting a framework specifically tailored for UAV-based detection tasks.
As for the overall process, let the real-world UAV-based detection dataset be denoted as $\mathcal{D}^{\text{real}}\!=\!\{(I_i^{\text{real}}, L_i^{\text{real}})\}_{i=1}^{N}$, consisting of $N$ image-annotation pairs. Each image $I_i^{\text{real}}$ is associated with an annotation set  $L_i^{\text{real}}\!=\!\{(b_j^{\text{real}},c_j^{\text{real}})\}_{j=1}^{n_i}$, which contains $n_i$ bounding boxes $b_j^{\text{real}}$ with their corresponding category labels $c_j^{\text{real}}\in\{1,2,...,C\}$.
A diffusion model $G_{\theta}(\cdot)$ with parameter $\theta$ is trained on $\mathcal{D}^{\text{real}}$ to synthesize a new images conditioned on the layout $L_i^{\text{real}}$ and auxiliary information $\mathcal{C}_i$, thereby forming a synthetic dataset $\mathcal{D}^{\text{syn}}$ as~\cref{eq:generate}.
\begin{equation}
    I_i^{\text{syn}} = G_{\theta}(L_i^{\text{real}},\mathcal{C}_i), ~~\mathcal{D}^{\text{syn}}\!=\!\{(I_i^{\text{syn}}, L_i^{\text{real}})\}_{i=1}^{M}.
    \label{eq:generate}
\end{equation}
Finally, $\mathcal{D}^{\text{syn}}$ and $\mathcal{D}^{\text{real}}$ are combined for detector training:
\begin{equation}
    \phi^* = \arg\min_{\phi} \mathcal{L}_{\text{det}}(F_{\phi}(\mathcal{D}^{\text{real}}\cup\mathcal{D}^{\text{syn}})),
    \label{eq:detector_train}
\end{equation}
where $F_{\phi}(\cdot)$ denotes the desired object detector with trainable parameter $\phi$ and $\mathcal{L}^{\text{det}}$ is the standard detection loss.

However, such general paradigm encounters challenges in UAV scenarios due to the previously discussed complex imaging conditions, including small scale, uneven spatial distribution and overlapping.
To address these issues, we introduce an enhanced data synthetic framework, namely UAVGen, which improves the fidelity and label consistency of $\mathcal{D}^{\text{syn}}$, thereby facilitating the optimization of~\cref{eq:detector_train}.
Specifically, we first introduce a Visual Prototype Conditioned Diffusion Model (VPC-DM) to produce more realistic UAV imagery.
Subsequently, a Focal Region-Enhanced Data Pipeline (FRE-DP) is designed to further synthesize the generated images and improve their consistency with corresponding labels. 

\subsection{Visual Prototype Conditioned Diffusion Model}
\label{sec:vpc-dm}
\noindent\textbf{Dual-Criterion Visual Prototype Selection.}
When training $G_\theta(\cdot)$ for conditioned generation, a common practice is directly cropping all object regions $L_i^{\text{real}}$ from $\mathcal{D}^{\text{real}}$ to serve as supervision. However, due to the small-scale appearance and dense distribution with overlap of UAV objects, these cropped patches are prone to suffer from noise and low quality, yielding unstable diffusion training.
To ensure reliable supervision, we instead employ high-quality target regions, referred to as visual prototypes, which are selected sequentially according to two criteria.

First, we retain prototypes with clear appearance and accurate localization in the visual space.
Taking advantage of the strong generalization capability, we use a pretrained detector $D(\cdot)$ to collect all detected regions from $\mathcal{D}^{\text{real}}$, and then group them by predicted class. This results in $\mathcal{G}^c=\{(b_i^{\text{det}}, s_i)\}_{i=1}^{K_c}$ for each class $c$, where $b_i^{\text{det}}$ denotes the detected bounding box and $s_i$ is its confidence score.
Noting that the confidence score for each class $c$ typically follows a normal distribution $\mathcal{N}(\mu_c,\sigma_c^2)$, we select high-confidence and well-localized candidates according to:
\begin{equation}
\begin{split}
    \mathcal{V}^c=\{b_i^{\text{real}}|&(b_i^{\text{det}}, s_i)\in \mathcal{G}^c,\\
    &\mathrm{IoU}(b_i^{\text{real}},b_i^{\text{det}})\ge \tau^{\text{det}}, s_i\ge \Phi^{-1}_c(\alpha)\},
    \label{eq:select_iou}
\end{split}
\end{equation}
where $\tau^{\text{det}}$ is the IoU threshold, and $\Phi^{-1}_c(\alpha)$ represents the $\alpha$-quantile of the confidence distribution for $\mathcal{G}^c$.

We further refine the candidates in the latent space to clarify fine-grained class boundaries and reduce generation confusion.
Concretely as shown in~\cref{eq:select_latent}, for each candidate region $b\in \mathcal{V}^c$, we extract the corresponding image patch and encode it using a VAE encoder $\mathcal{E}_{\text{img}}(\cdot)$ to obtain a latent representation. Then, the visual prototype set $\mathcal{P}^c$ for class $c$ is constructed by selecting candidates whose latent embeddings lie close to the class-specific centroid $\bm{\mu}^c$.
\begin{equation}
    \begin{split}
        \mathcal{P}^c=\{b\in\mathcal{V}^c\mid \left\| \mathcal{E}_{\text{img}}(b)-\bm{\mu}^c \right\|^2<\tau^{\text{lat}}\}, \\
        \bm{\mu}^c = \frac{1}{|\mathcal{V}^c|} \sum_{b\in\mathcal{V}^c} \mathcal{E}_{\text{img}}(b).~~~~~~~~~~~~~
    \end{split}
    \label{eq:select_latent}
\end{equation}

\noindent\textbf{Multi-Source Condition Encoding.}
To achieve controllable and semantically consistent generation, we design a Multi-Source Condition Encoding mechanism that provides complementary conditioning from both visual-prototype-enhanced layouts and textual information.

Given a layout $L_i^{\text{real}}$ as defined in~\cref{sec:framework_overview}, for each region $(b_j^{\text{real}},c_j^{\text{real}})\in L_i^{\text{real}}$, we sample a visual prototype $P_j$ from the corresponding class-specific set $\mathcal{P}^c$, where $c=c_j^{\text{real}}$.
Then, $P_j$ is placed at position $b_{j}^{\text{real}}$ with geometrical transformation to match the target scale, forming a composite image $I_j^{\text{blank}}$ on a blank (\ie~zero-valued) canvas.
Finally, we obtain a layout embedding $\bm{v}_i$, which is enhanced by visual prototype, by aggregating the latent representations of all regions through a 3D convolutional network $\text{Conv}(\cdot)$ formally expressed as~\cref{eq:layout_embedding}. The module fuses features along the prototype dimension: $N$ features of shape $B \times C \times H \times W$ are stacked to $B \times N \times C \times H \times W$, fused to $B \times 1 \times C \times H \times W$. Learnable padding handles cases with fewer than $N$ prototypes.
\begin{equation}
    \bm{v}_{i}\!=\!\text{Conv} (\mathcal{E}_\text{img}(I_1^{\text{blank}}), \mathcal{E}_\text{img}(I_2^{\text{blank}}), ..., \mathcal{E}_\text{img}(I_{n_i}^{\text{blank}}) ).
    \label{eq:layout_embedding}
\end{equation}

Furthermore, to complement visual cues with class semantics, we construct layout-aware textual embeddings.
Specifically, we introduce two textual prompts describing the overall scene and individual objects, respectively:
\begin{equation}
    \begin{split}
        t_i^g\!&=\!\text{`}An~aerial~image~with~ \{c_1\},\{c_2 \},...,\{c_{n_i}\}.\text{'}\\
        t^{c_j}\!&=\!\text{`}An~aerial~image~of~\{c_j\}.\text{'}
    \end{split}
\end{equation}
As shown in~\cref{eq:text_condition}, $t_i^g$ is directly encoded by a text encoder $\mathcal{E}_{\text{text}}(\cdot)$ to gain the global embedding $\bm{e}_i^g$.
To capture fine-grained, object-level semantics, we enrich each object embedding with positional information via Fourier embeddings $\mathcal{F}(\cdot)$, inspired by GLIGEN \cite{li2023gligen}.
Consequently, each object-specific text embedding is concatenated with its corresponding positional embedding and processed through an MLP-based network, after which all $n_i$ features are aggregated by a gated attention network ${GA}(\cdot)$, resulting in a unified fine-grained layout embedding $\bm{e}_i^f$.
\begin{equation}
\begin{split}
    \bm{e}_i^g\!=\!\mathcal{E}_{\text{text}}(t^g),~~~~~~~~~~~~~~~~~~~~~~~~~~~~~~\\
    \bm{e}_i^f\!=\!{GA}\!\left(\left\{\mathrm{MLP}([\mathcal{E}_{\text{text}}( t^{c_j});\mathcal{F}(b_j^{\text{real}})])\right\}_{j=1}^{n_i}\right).
\end{split}
\label{eq:text_condition}
\end{equation}

\noindent\textbf{Condition Injection and Training.}
Building upon the standard diffusion process parameterized by the noisy latent $\bm{x}_t$ and timestep $t$, we incorporate the previously obtained multi-source conditions to enhance controllable and layout-consistent generation.
Specifically, the visual prototype-based layout embedding $\bm{v}_i$ and the fine-grained textual embedding $\bm{e}_i^f$ are injected through a ControlNet ${CN}(\cdot )$, which produces a layout-aware conditional feature $\bm{\mathcal{C}}_i$ as:
\begin{equation}
    \bm{\mathcal{C}}_i = {CN}(\bm{x}_t, t\mid \bm{v}_i, \bm{e}_i^f).
\end{equation}
Denoting $\epsilon_{\theta}$ as the noise prediction network of $G_{\theta}$, both $\bm{e}_i^g$ and $\bm{\mathcal{C}}_i$ jointly guide the denoising process by $\epsilon_{\theta}$. This enables the generative process to maintain the overall scene semantics as well as preserving fine-grained object layouts.

To further enhance the fidelity of target objects, we employ a foreground-aware reweighted loss that emphasizes supervision on object regions:
\begin{equation}
    \mathcal{L}_{\text{layout}}\!=\!\bm{w} \odot \mathbb{E}_{\bm{x}_0, t, \epsilon \sim \mathcal{N}(\bm{0}, \bm{1})} \left[ \| \epsilon\!-\!\epsilon_\theta(\bm{x}_t, t | \bm{e}_i^g, \mathcal{C}_i) \|^2 \right],
\end{equation}
where $\bm{w}$ is a spatial weighting map derived from $L_i^{\text{real}}$. Pixels within object regions are assigned a higher weight (\ie~$>1$), while background regions retain a unit weight.

\subsection{Focal Region Enhanced Data Pipeline.}
\noindent\textbf{Region-based Data Synthesis.} 
UAV-based images typically exhibit small-object clustering, with the majority of image regions containing few or even no target instances. Such sparsely populated images may both limit the effectiveness of diffusion-based generation and reduce the training efficiency of object detection.
To address this issue, we concentrate on object-rich areas, termed focal regions, to enhance both data synthesis and detector training.

Given a layout $L_i^{\text{real}}$ as defined in~\cref{sec:framework_overview}, we first compute the geometric center of each $b_j^{\text{real}}$, denoted as $\bm{p}_j$. A K-means clustering is then applied to the set of centroids $\{p_j\}_{j=1}^{n_i}$ to obtain $K$ cluster centers $\{\bm{m}_k\}_{k=1}^K$ such that:
\begin{equation}
    \bm{m}_k=\arg\min_{\bm{m}}\sum_{\bm{p}_j\in \mathcal{S}_k}\left\|\bm{p}_j-\bm{m}\right\|^2,
\end{equation}
where $\mathcal{S}_k$ is the set of centroids assigned to cluster $k$.

Next, for each $\bm{m}_k$, the corresponding focal region $\mathcal{B}_k$ is determined by solving an overlap maximization problem:
\begin{equation}
    \mathcal{B}_k = \arg\max_{\mathcal{B}\in\Omega(\bm{m}_k)}\sum_{j=1}^{n_i}\mathbb{I}\left[b_j^{\text{real}}\cap \mathcal{B}=b_j^{\text{real}}\right],
\end{equation}
where $\Omega(\bm{m}_k)$ is the set of candidate regions with identical size containing $\bm{m}_k$.

The selected object-dense regions are then cropped to construct a high-information-density dataset $\mathcal{D}^{\text{real\_dense}}$, which serves as a refined substitute for $\mathcal{D}^{\text{real}}$ and is subsequently used as the training and generation target for our proposed VPC-DM (\ie~\cref{sec:vpc-dm}).

Finally, the generated focal-region images are merged back into the original image resolution to form a high-information-density synthetic dataset $\mathcal{D}^{\text{syn\_dense}}$ as an alternative of the typical $\mathcal{D}^{\text{syn}}$. This dataset is then used to train the detector as formulated in~\cref{eq:detector_train}.

\noindent\textbf{Label Refinement.}
Despite advances by our proposed layout-to-image synthesis framework, discrepancies between synthesized images and input layouts remain a significant challenge, especially for UAV imagery characterized by small-scale objects. We categorize these inconsistencies into three types: \textit{missed generations}, \textit{false generations}, and \textit{label misalignments}. To mitigate these issues, we propose a post-processing pipeline with label refinement.

Given a synthesized sample $(I^{\text{syn}},L^{\text{real}})\in\mathcal{D}^{\text{syn\_dense}}$ obtained above (for simplicity, the subscript `$^{\text{dense}}$' is omitted), we apply a pretrained detector $D(\cdot)$ on $I^{\text{syn}}$, obtaining predicted results:
\begin{equation}
    L^{\text{det}}=D(I^{\text{syn}})=\{(b_j^{\text{det}},c_j^{\text{det}})\}_{j=1}^{N}.
    \label{eq:label_detect}
\end{equation}
We then perform IoU-based matching with threshold $\tau^{\text{ref}}$ between predicted boxes and real boxes, resulting in:
\begin{equation}
    \mathcal{M}=\{(b_i^{\text{real}}, b_j^{\text{det}})\mid \mathrm{IoU}(b_i^{\text{real}}, b_j^{\text{det}})\ge\tau^{\text{ref}}\}.
    \label{eq:pairs}
\end{equation}

First, missed generations refer to regions where the diffusion model fails to generate the corresponding object, denoted as $L^{\text{miss}}$.
To discard missing instances from labels and filter out low-quality ones, we preserve the labels aligned with detection results $L^{\text{det}}$ and discard low-quality ones with threshold $\alpha$ on corresponding detection confidence scores. The refined label set is updated as~\cref{eq:refine1}.
\begin{equation}
\begin{split}
    L^{\text{miss}}=L^{\text{real}} \setminus \{(b_i^{\text{real}},c_i^{\text{real}})\mid(b_i^{\text{real}}, b_j^{\text{det}})\in \mathcal{M}\}, ~~\\
    L^{\text{ref}} = \{(b_i^{\text{real}}, c_i^{\text{real}}) \in L^{\text{real}} \setminus L^{\text{miss}} \mid s_j\ge \Phi_{c_j}^{-1}(\alpha) \}
\end{split}
\label{eq:refine1}
\end{equation}

Second, false generations represent extra generated objects, denoted as $L^{\text{false}}$. Thus, we apply a confidence filter per-class and update the refined set as~\cref{eq:refine2}.
\begin{equation}
\begin{split}
    L^{\text{false}}= L^{\text{det}} \setminus \{(b_j^{\text{det}},c_j)\mid(b_i^{\text{real}}, b_j^{\text{det}})\in \mathcal{M}\},~~~ \\
    L^{\text{ref}}\leftarrow L^{\text{ref}}\cup \{(b_j^{\text{det}},c_j^{\text{det}})\in L^{\text{false}} \!\mid\! s_j\ge \Phi_{c_j}^{-1}(\beta)\}.
\end{split}
\label{eq:refine2}
\end{equation}

Third, for label misalignments, we focus on mitigating label discrepancies by refining each $(b_j^{\text{ref}}, c_j^{\text{ref}})\in L^{\text{ref}}$ as~\cref{eq:refine3} with $\gamma$ controlling reliance on predictions.
\begin{equation}
(b_j^{\text{ref}}, c_j^{\text{ref}})\leftarrow
\begin{cases} 
    (b_j^{\text{det}}, c_j^{\text{det}}) & \text{if } s_j > \Phi_{c_j}^{-1}(\gamma) \\
    (b_j^{\text{ref}}, c_j^{\text{ref}}) & \text{otherwise}
\end{cases}.
\label{eq:refine3}
\end{equation}

Ultimately, the refined dataset $\mathcal{D}^{\text{ref}}=(I^{\text{syn}},L^{\text{ref}})$ is employed to construct the training set for object detector. $\alpha$, $\beta$, and $\gamma$ are set according to the accuracy of detector $D(\cdot)$.

\begin{table*}[!t]
\centering
\caption{Comparison of FID scores and Average Precision (\%) of UAV-based detector trained with augmented data produced by different methods on VisDrone and UAVDT. “Real only” means the detector is trained only on real data. The best results are highlighted in \textbf{bold}.}
\label{table1}
\begin{tabular}{c l c ccc ccc}
\toprule
    \multirow{2}{*}{\textbf{Dataset}} & \multirow{2}{*}{\textbf{Method}} & \multirow{2}{*}{\textbf{FID$\downarrow$}} & \multicolumn{6}{c}{\textbf{Average Precision$\uparrow$}} \\
    \cmidrule(){4-9}
    & & & mAP & AP$_{50}$ & AP$_{75}$ & AP$_{s}$ & AP$_{m}$ & AP$_{l}$ \\
\midrule
    \multirow{6}{*}{VisDrone \cite{du2019visdrone}} & Real only & - & 24.5 & 42.1 & 24.5 & 15.4 & 36.1 & 37.9 \\
\cmidrule(l){2-9}
    & CopyPaste \cite{dwibedi2017cut} & - & 23.6 & 41.0 & 23.5 & 14.7 & 34.6 & 40.2 \\
\cmidrule(l){2-9}
    & GLIGEN \cite{li2023gligen} & 72.68 & 24.8 & 43.0 & 24.6 & 15.6 & 36.3 & 39.0 \\
    & Geodiffusion \cite{chen2024geodiffusion} & 57.96 & 24.7 & 42.9 & 24.7 & 15.4 & 36.1 & 39.1 \\
    & AeroGen \cite{tang2025aerogen} & 48.04 & 24.9 & 43.3 & 24.5 & 15.6 & 36.3 & 43.3 \\
    & \textbf{UAVGen (Ours)} & \textbf{34.34} & \textbf{25.9} & \textbf{44.8} & \textbf{26.0} & \textbf{16.7} & \textbf{37.5} & \textbf{43.7} \\
\midrule
    \multirow{5}{*}{UAVDT \cite{du2018unmanned}} & Real only & - & 14.5 & 26.1 & 14.7 & 10.3 & 24.1 & 11.8 \\
\cmidrule(l){2-9}
    & Geodiffusion \cite{chen2024geodiffusion} & 47.81 & 14.0 & 28.4 & 12.1 & 10.3 & 24.2 & 12.2 \\
    & AeroGen \cite{tang2025aerogen} & 31.99 & 15.0 & 28.3 & 14.5 & 10.7 & 25.1 & 12.6 \\
    & \textbf{UAVGen (Ours)} & \textbf{29.73} & \textbf{16.6} & \textbf{30.9} & \textbf{16.1} & \textbf{11.7} & \textbf{27.8} & \textbf{14.8} \\
\bottomrule
\end{tabular}
\end{table*}
\section{Experimental Results and Analysis}
\label{sec:experiment}

\begin{figure}[t]
\centering
\includegraphics[width=0.9 \columnwidth]{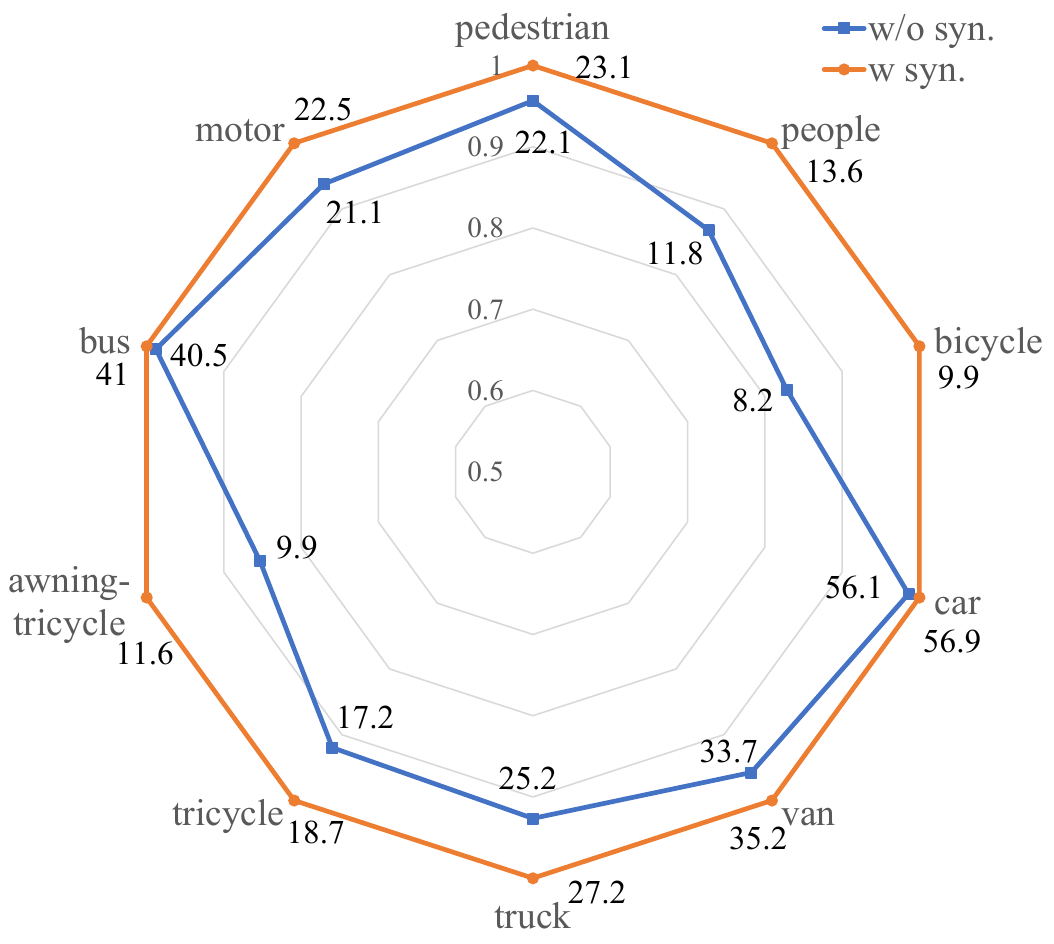}
\caption{Comparison of mAP across different categories on VisDrone. Our method (w syn.) generates 738 synthesized images, yields consistent performance improvements across all categories when compared to the non-augmented baseline (w/o syn.).}
\label{FigRadar}
\end{figure}

\subsection{Experimental Settings}
\noindent\textbf{Dataset and Evaluation Metrics.}
We evaluate our proposed UAVGen on two public UAV object detection datasets. VisDrone~\cite{du2019visdrone} is a UAV-view benchmark with 6,471 training, 548 validation, and 1,580 test images, covering 10 categories including pedestrians and vehicles. UAVDT~\cite{du2018unmanned} is a UAV detection and tracking benchmark, comprising 24,143 training and 16,592 test images sampled from UAV-captured videos. Both datasets pose core challenges for UAV-based object detection, including abundant small objects with extreme scale variations, dense targets, frequent occlusions, and dynamic scenes.

We adopt Fréchet Inception Distance (FID) \cite{heusel2017gans} and Average Precision (AP) \cite{everingham2010pascal} as our evaluation metrics, where FID measures image generation quality and AP assesses object detection accuracy. For AP, we evaluate mAP, AP$_{50}$, AP$_{75}$, AP${_s}$, AP${_m}$ and AP${_l}$, following the common size division criteria of MS COCO benckmark \cite{lin2014microsoft}.

\begin{figure*}[!t]
\centering
\includegraphics[width=0.90\textwidth]{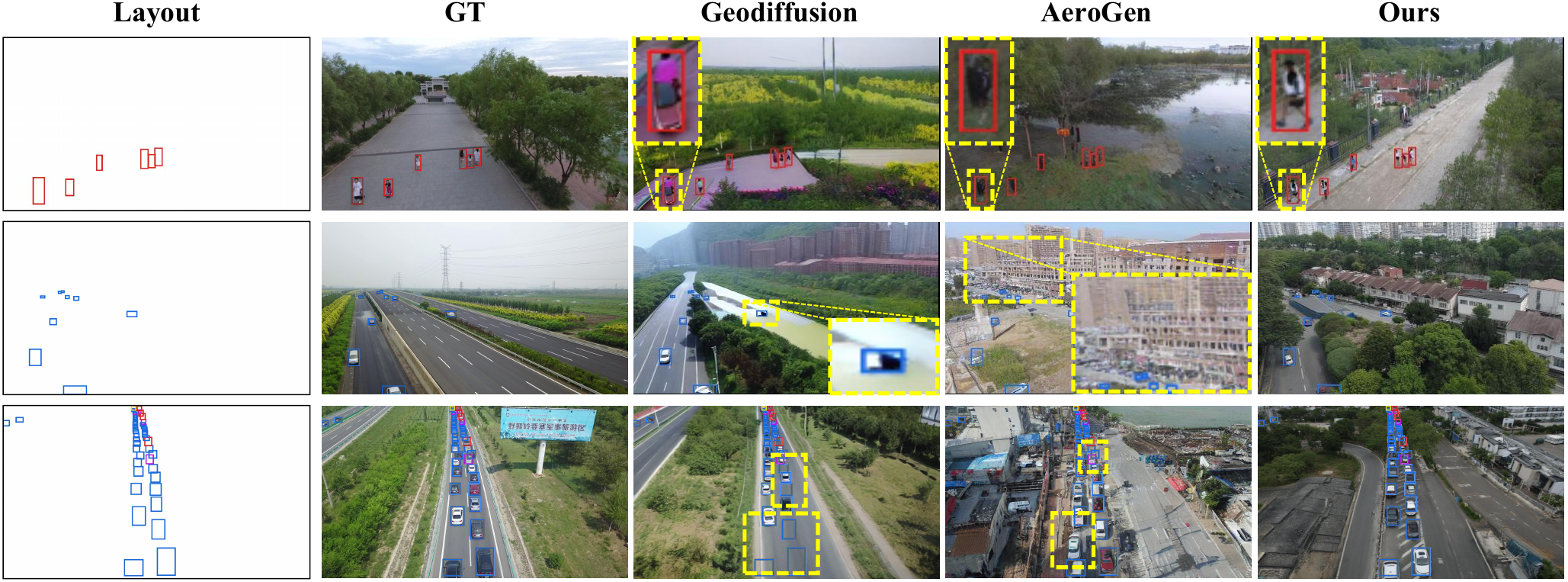}
\caption{Comparison of generated images on VisDrone. Our method exhibits superior layout-image consistency and enhanced visual fidelity of generated small-scale objects. The yellow dashed boxes denote the regions where the generated targets are inconsistent with the inputs, as well as blurry low-quality targets. For the generation of small pedestrian targets(line 1), our method achieves significantly higher clarity than the comparative methods. Furthermore, in scenarios with dense small targets(line 2, 3), our method exhibits superior consistency between objects and layout.}
\label{FigGenImg}
\end{figure*}

\noindent\textbf{Implementation Details.}
Our diffusion-driven data synthesis method is built on the FLUX model, trained with a learning rate of 1e-5 for 60K iterations, batch size 8, on one NVIDIA A800 GPU. Training and generation are performed at 512$\times$512 resolution. The diffusion model, text encoder, and VAE encoder are initialized from pretrained FLUX weights and frozen, with remaining parameters finetuned. We adopt Faster R-CNN~\cite{ren2016faster} as the base detector for visual prototype selection and label refinement. For fidelity evaluation, we generate an equal number of images to real data for FID computation; to ensure fair comparison, only VPC-DM is used here, while the full method is adopted for trainability evaluation. On VisDrone, we generate 18,649 patches and merge into 738 full images; on UAVDT, we generate 29,403 patches and merge into 9,802 full images. All compared methods follow standard practice, generating datasets matching the training set size (6,474 for VisDrone, 24,143 for UAVDT). We also reproduce the Copy-Paste baseline~\cite{dwibedi2017cut}, with masks inferred via DeepLabV3 ResNet101, synthesizing 7,258 images mixed with real data for GFL detector training on VisDrone.

\begin{table}[!t]
\centering
\caption{Comparison of Average Precisions (\%) achieved by different methods in enhancing the state-of-the-art UAV-based object detection model RemDet-X on VisDrone. The best results are highlighted in \textbf{bold}.}
\label{table2}
\begin{tabular}{l c c c}
\toprule
    \textbf{Method} & \textbf{{mAP}$\uparrow$} & \textbf{AP$_{50}$$\uparrow$} & \textbf{AP$_{75}$$\uparrow$} \\
\midrule
    Real only & 29.8 & 48.1 & 30.8 \\
\midrule
    GLIGEN \cite{li2023gligen} & 29.4 & 47.7 & 30.2 \\
    Geodiffusion \cite{chen2024geodiffusion} & 29.4 & 47.8 & 30.2 \\
    AeroGen \cite{tang2025aerogen} & 28.8 & 46.9 & 29.6 \\
    \textbf{UAVGen (Ours)} & \textbf{30.2} & \textbf{48.7} & \textbf{31.1} \\
\bottomrule
\end{tabular}
\end{table}

\subsection{Fidelity}
We evaluated fidelity of UAVGen on the VisDrone~\cite{du2019visdrone} dataset and the UAVDT~\cite{du2018unmanned} dataset, by comparing with the latest approaches including GLIGEN \citep{li2023gligen}, Geodiffusion \citep{chen2024geodiffusion} and AeroGEN~\citep{tang2025aerogen}. As represented in Table~\ref{table1}, UAVGen substantially outperforms the compared methods, reducing FID of UAV-based images generation by 13.7 and 2.3 on VisDrone and UAVDT, respectively. This indicates that by leveraging high-quality visual prototypes and fine-grained layout semantics, UAVGen effectively mitigates the challenges posed by small, overlapped objects in UAV imagery, leading to significantly improved generation quality.

We further visualized the synthetic results, as shown in Fig.\ref{FigGenImg}. UAVGen demonstrates strong consistency with the input layouts, particularly in challenging scenarios involving small and densely clustered objects. Our focal-region generation strategy effectively improves the generation quality of small objects, producing them with noticeably higher fidelity compared to existing methods.

\subsection{Trainability}
\noindent\textbf{General Detector Enhancement.} For UAV-based object detection, we adopt GFL \citep{li2020generalized} as the representative general-purpose model. We trained GFL-ResNet50 for 12 epochs.
As shown in Table~\ref{table1}, on VisDrone benchmark, UAVGen achieves state-of-the-art performance with only 738 augmented images, yielding 1.4\% improvement in mAP over baseline methods. UAVGen achieves 1.3\% improvement in AP$_s$, while the remaining methods show no improvement in small object detection. Furthermore, our approach attains superior mAP across all categories on VisDrone benchmark, with mAP of several categories shown in Fig.~\ref{FigRadar}. 
On UAVDT benckmark, UAVGen yields 2.1\% improvement in mAP and 4.8\% improvement in AP$_{50}$. Compared to Geodiffusion \cite{chen2024geodiffusion}, our method achieves measurable gains in small-target detection precision, with 1.5\% improvement on AP$_s$.

\noindent\textbf{SOTA Detector Enhancement.}
Beyond conventional detection models, we conducted experiment on RemDet \cite{li2025remdet}, the state-of-the-art detector based on YOLO for the VisDrone dataset. RemDet was trained using default configurations. As shown in Table~\ref{table2}, UAVGen delivers comprehensive performance gains in mAP when applied to RemDet. Notably, comparative methods exhibit detrimental effects on SOTA detector. The experiment demonstrates the efficacy of UAVGen for UAV-based object detection.

\subsection{Ablation Study}
\begin{table}[!t]
\centering
\caption{Ablation study on proposed methods in generation pipeline. "Gen." suggests train detector with generated images, “VP” suggests Visual Prototype, "LE" suggests Layout Embeddings, "FR" suggests Focal Region Generation, "LR" suggests Label Refinement.}
\label{table4}
\begin{tabular}{ccc cc cc}
\toprule
    \multicolumn{3}{c}{\textbf{VPC-DM}} & \multicolumn{2}{c}{\textbf{FRE-DP}} & \multirow{2}{*}{\textbf{mAP$\uparrow$}} & \multirow{2}{*}{\textbf{AP$_{50}$$\uparrow$}} \\
\cmidrule(r){1-3} \cmidrule(l){4-5}
    Gen. & VP & LE & FR & LR &  & \\
\midrule
    & & & & & 24.5 & 42.1 \\
    \Checkmark & & & & & 23.8 & 42.1 \\
    \Checkmark & \Checkmark & & & & 24.1 & 42.4 \\
    \Checkmark & \Checkmark & \Checkmark & & & 25.2 & 43.8 \\
    \Checkmark & \Checkmark & \Checkmark & \Checkmark & & 25.5 & 44.5 \\
    \Checkmark & \Checkmark & \Checkmark & & \Checkmark & 25.5 & 44.2 \\
\midrule
    \Checkmark & \Checkmark & \Checkmark & \Checkmark & \Checkmark & \textbf{25.9} & \textbf{44.8} \\
\bottomrule
\end{tabular}
\end{table}

\begin{table}[!t]
\caption{Ablation study on resolution scale of focal region.}
\label{table5}
\centering
\begin{tabular}{c ccc}
\toprule
    \textbf{FR Res.} & \textbf{mAP$\uparrow$} & \textbf{AP$_{50}$$\uparrow$} & \textbf{AP$_{75}$$\uparrow$} \\
\midrule
    1024 & 25.3 & 43.9 & 25.3 \\
    512 & 25.6 & 44.7 & 25.4 \\
    256 & \textbf{25.9} & \textbf{44.8} & \textbf{26.0} \\
\bottomrule
\end{tabular}
\end{table}

\noindent\textbf{On VPC-DM.} 
To validate the effectiveness of the data generated by VPC-DM, we conducted ablation experiments on Visual Prototypes and Layout Embeddings. As shown in in Table~\ref{table4}, when both components are employed, the detector’s performance improves from 24.5\% to 25.2\%. Notably, the generated images are detrimental to detector training when neither component is used, while the beneficial impact of the generated data on detector gradually strengthens as the two components are incorporated incrementally.

\noindent\textbf{On FRE-DP.}
We conducted ablation study on the Focal Region Generation Strategy and Label Refinement, with results summarized in Table~\ref{table4}. Specifically, based on VPC-DM, FRE-DP improves the baseline mAP from 25.2\% to 25.9\%, indicating the effectiveness of enhancing detection performance. Moreover, both VPC-DM and FRE-DP enhance the quality of the generated images, improving the performance of detector (each brings +0.7\%). When both applied concurrently, detector achieves peak mAP 25.9\%.

\begin{figure}[t]
\centering
\includegraphics[width=0.95 \columnwidth]{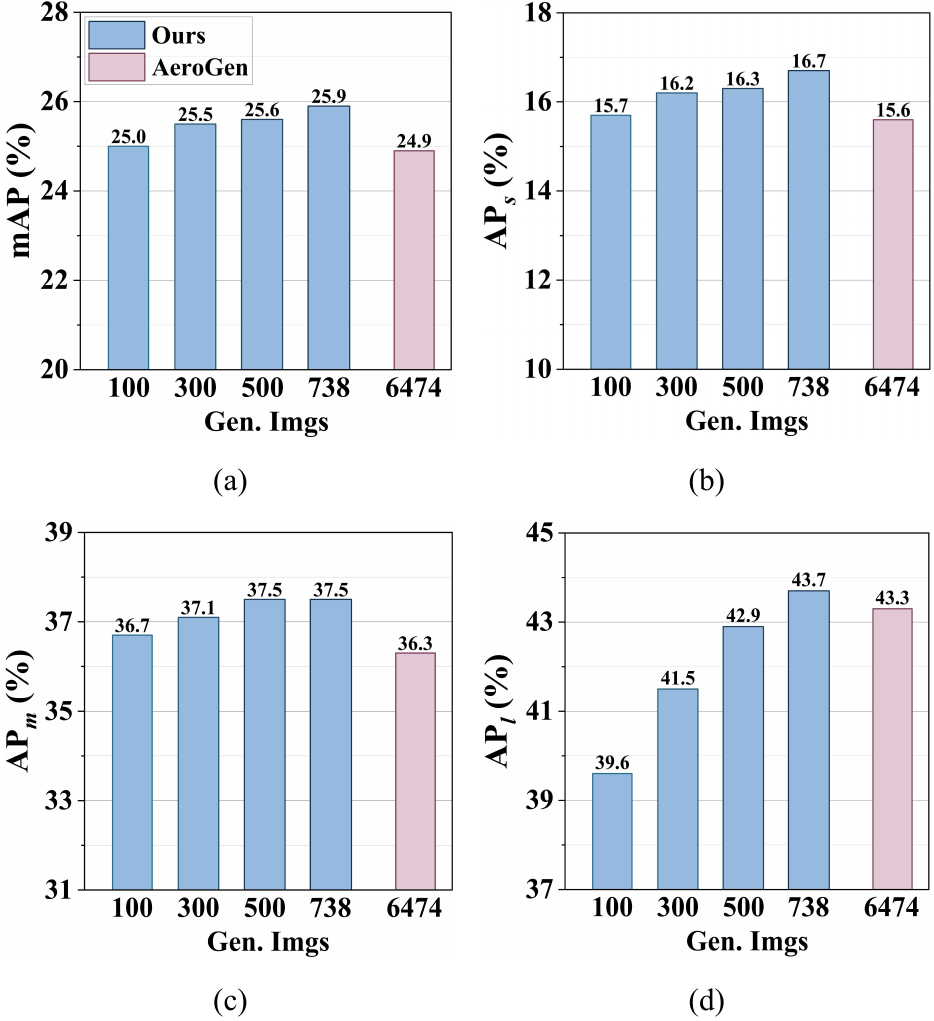}
\caption{Impact of various scale of generated images on object detection model.  With reduced images, our method achieves superior performance.} 
\label{FigGenScale}
\end{figure}

\noindent\textbf{On Synthesized Data Scale.}
To validate the effectiveness of synthetic images generated by UAVGen, we trained GFL using randomly sampled sets of 100, 300, 500, and 738 (all) synthesized images. As shown in Fig.\ref{FigGenScale}, the detector with 100 additional training images achieved performance comparable to AeroGen, which utilizes 6,474 (all) images. The performance of detector exhibits consistent improvement as the number of synthetic training images increases, confirming the effectiveness of our method. Furthermore, our method achieves superior enhancement with fewer synthesized images, demonstrating the efficiency of FRE-DP.

\noindent\textbf{On Region Scale.}
We also performed an ablation study on resolution within the Focal Region Generation, as summarized in Table~\ref{table5}. We evaluated effect of focal region at different resolutions: 1024, 512, and 256. The results demonstrate a clear trend: as the region resolution decreases, the quality of the generated object regions improves, leading to enhanced detection performance. When using the smallest resolution of 256, the detector achieves the best results, with mAP 25.9\%. This indicates that lower region resolutions can more effectively enhance the quality of small object images and contribute to superior model performance. 
\section{Conclusion}
\label{sec:conclusion}
In this paper, we introduce UAVGen, a diffusion-driven data augmentation framework for UAV-based object detection: we construct high-quality layout conditions for image generation via our Visual Prototype Conditioned Diffusion Model (VPC-DM), focus synthesis on focal regions, and correct misalignments through our Focal Region Enhanced Data Pipeline (FRE-DP), fully exploiting generative potential and enhancing synthesized data consistency. Extensive experiments on diverse datasets demonstrate the effectiveness of our method, which achieves superior image generation fidelity and significantly improves UAV-based detector performance. While tailored for dense small-object detection in UAV scenes, our general-purpose fidelity enhancement and label–image alignment modules extend naturally to general object detection, and FRE-DP’s small-object optimization brings notable benefits to far-distant surveillance and remote sensing scenarios, with validation planned for future work. Addressing appearance and scale variations from viewpoint and altitude changes in real-world UAV scenarios remains a promising avenue for further research.

\clearpage
\section*{Acknowledgments}
\label{sec:acknowledgments}
This work was partly supported by the New Generation Artificial Intelligence-National Science and Technology Major Project (2025ZD0124000), the Beijing Natural Science Foundation (No. 4242044 and No. L259044), the CCF Baidu Open Fund, and the Fundamental Research Funds for the Central Universities.
{
    \small
    \bibliographystyle{ieeenat_fullname}
    \bibliography{main}

\begin{thebibliography}{62}
\providecommand{\natexlab}[1]{#1}
\providecommand{\url}[1]{\texttt{#1}}
\expandafter\ifx\csname urlstyle\endcsname\relax
  \providecommand{\doi}[1]{doi: #1}\else
  \providecommand{\doi}{doi: \begingroup \urlstyle{rm}\Url}\fi

\bibitem[Bar-Tal et~al.(2023)Bar-Tal, Yariv, Lipman, and Dekel]{bar2023multidiffusion}
Omer Bar-Tal, Lior Yariv, Yaron Lipman, and Tali Dekel.
\newblock Multidiffusion: Fusing diffusion paths for controlled image generation.
\newblock In \emph{Proceedings of the International Conference on Machine Learning}, pages 1737--1752, 2023.

\bibitem[Brock et~al.(2018)Brock, Donahue, and Simonyan]{brock2018large}
Andrew Brock, Jeff Donahue, and Karen Simonyan.
\newblock Large scale gan training for high fidelity natural image synthesis.
\newblock In \emph{Proceedings of the International Conference on Learning Representations}, pages 9256--9291, 2018.

\bibitem[Chen et~al.(2024)Chen, Xie, Chen, Wang, Hong, Li, and Yeung]{chen2024geodiffusion}
Kai Chen, Enze Xie, Zhe Chen, Yibo Wang, Lanqing Hong, Zhenguo Li, and Dit-Yan Yeung.
\newblock Geodiffusion: Text-prompted geometric control for object detection data generation.
\newblock In \emph{Proceedings of the International Conference on Learning Representations}, pages 846--868, 2024.

\bibitem[Cheng et~al.(2023)Cheng, Liang, Shi, He, Xiao, and Li]{cheng2023layoutdiffuse}
Jiaxin Cheng, Xiao Liang, Xingjian Shi, Tong He, Tianjun Xiao, and Mu Li.
\newblock Layoutdiffuse: Adapting foundational diffusion models for layout-to-image generation.
\newblock \emph{arXiv preprint arXiv:2302.08908}, 2023.

\bibitem[Dhariwal and Nichol(2025)]{dhariwal2021diffusion}
Prafulla Dhariwal and Alexander Nichol.
\newblock Diffusion models beat gans on image synthesis.
\newblock In \emph{Advances in Neural Information Processing Systems}, pages 4643--4651, 2025.

\bibitem[Dieter et~al.(2023)Dieter, Weinmann, J{\"a}ger, and Brucherseifer]{dieter2023quantifying}
Tamara~Regina Dieter, Andreas Weinmann, Stefan J{\"a}ger, and Eva Brucherseifer.
\newblock Quantifying the simulation--reality gap for deep learning-based drone detection.
\newblock \emph{Electronics}, 12\penalty0 (10):\penalty0 2197, 2023.

\bibitem[Du et~al.(2023)Du, Huang, Chen, and Huang]{du2023adaptive}
Bowei Du, Yecheng Huang, Jiaxin Chen, and Di Huang.
\newblock Adaptive sparse convolutional networks with global context enhancement for faster object detection on drone images.
\newblock In \emph{Proceedings of the IEEE/CVF Conference on Computer Vision and Pattern Recognition}, pages 13435--13444, 2023.

\bibitem[Du et~al.(2018)Du, Qi, Yu, Yang, Duan, Li, Zhang, Huang, and Tian]{du2018unmanned}
Dawei Du, Yuankai Qi, Hongyang Yu, Yifan Yang, Kaiwen Duan, Guorong Li, Weigang Zhang, Qingming Huang, and Qi Tian.
\newblock The unmanned aerial vehicle benchmark: Object detection and tracking.
\newblock In \emph{Proceedings of the European Conference on Computer Vision}, pages 370--386, 2018.

\bibitem[Du et~al.(2019)Du, Zhu, Wen, Bian, Lin, Hu, Peng, Zheng, Wang, Zhang, et~al.]{du2019visdrone}
Dawei Du, Pengfei Zhu, Longyin Wen, Xiao Bian, Haibin Lin, Qinghua Hu, Tao Peng, Jiayu Zheng, Xinyao Wang, Yue Zhang, et~al.
\newblock Visdrone-det2019: The vision meets drone object detection in image challenge results.
\newblock In \emph{Proceedings of the IEEE/CVF International Conference on Computer Vision Workshops}, pages 213--226, 2019.

\bibitem[Dvornik et~al.(2018)Dvornik, Mairal, and Schmid]{dvornik2018modeling}
Nikita Dvornik, Julien Mairal, and Cordelia Schmid.
\newblock Modeling visual context is key to augmenting object detection datasets.
\newblock In \emph{Proceedings of the European Conference on Computer Vision}, pages 364--380, 2018.

\bibitem[Dwibedi et~al.(2017)Dwibedi, Misra, and Hebert]{dwibedi2017cut}
Debidatta Dwibedi, Ishan Misra, and Martial Hebert.
\newblock Cut, paste and learn: Surprisingly easy synthesis for instance detection.
\newblock In \emph{Proceedings of the IEEE/CVF International Conference on Computer Vision}, pages 1301--1310, 2017.

\bibitem[Erdelj et~al.(2017)Erdelj, Natalizio, Chowdhury, and Akyildiz]{erdelj2017help}
Milan Erdelj, Enrico Natalizio, Kaushik~R Chowdhury, and Ian~F Akyildiz.
\newblock Help from the sky: Leveraging uavs for disaster management.
\newblock \emph{IEEE Pervasive Computing}, 16\penalty0 (1):\penalty0 24--32, 2017.

\bibitem[Everingham et~al.(2010)Everingham, Van~Gool, Williams, Winn, and Zisserman]{everingham2010pascal}
Mark Everingham, Luc Van~Gool, Christopher~KI Williams, John Winn, and Andrew Zisserman.
\newblock The pascal visual object classes (voc) challenge.
\newblock \emph{International Journal of Computer Vision}, 88\penalty0 (2):\penalty0 303--338, 2010.

\bibitem[Gao et~al.(2024)Gao, Chen, Xie, HONG, Li, Yeung, and Xu]{gao2023magicdrive}
Ruiyuan Gao, Kai Chen, Enze Xie, Lanqing HONG, Zhenguo Li, Dit-Yan Yeung, and Qiang Xu.
\newblock Magicdrive: Street view generation with diverse 3d geometry control.
\newblock In \emph{Proceedings of the International Conference on Learning Representations}, pages 904--923, 2024.

\bibitem[Ghiasi et~al.(2021)Ghiasi, Cui, Srinivas, Qian, Lin, Cubuk, Le, and Zoph]{ghiasi2021simple}
Golnaz Ghiasi, Yin Cui, Aravind Srinivas, Rui Qian, Tsung-Yi Lin, Ekin~D Cubuk, Quoc~V Le, and Barret Zoph.
\newblock Simple copy-paste is a strong data augmentation method for instance segmentation.
\newblock In \emph{Proceedings of the IEEE/CVF Conference on Computer Vision and Pattern Recognition}, pages 2918--2928, 2021.

\bibitem[Goodfellow et~al.(2014)Goodfellow, Pouget-Abadie, Mirza, Xu, Warde-Farley, Ozair, Courville, and Bengio]{goodfellow2014generative}
Ian~J Goodfellow, Jean Pouget-Abadie, Mehdi Mirza, Bing Xu, David Warde-Farley, Sherjil Ozair, Aaron Courville, and Yoshua Bengio.
\newblock Generative adversarial nets.
\newblock In \emph{Advances in Neural Information Processing Systems}, pages 2672--2680, 2014.

\bibitem[Heusel et~al.(2017)Heusel, Ramsauer, Unterthiner, Nessler, and Hochreiter]{heusel2017gans}
Martin Heusel, Hubert Ramsauer, Thomas Unterthiner, Bernhard Nessler, and Sepp Hochreiter.
\newblock Gans trained by a two time-scale update rule converge to a local nash equilibrium.
\newblock In \emph{Advances in Neural Information Processing Systems}, page 6629–6640, 2017.

\bibitem[Ho et~al.(2020)Ho, Jain, and Abbeel]{ho2020denoising}
Jonathan Ho, Ajay Jain, and Pieter Abbeel.
\newblock Denoising diffusion probabilistic models.
\newblock In \emph{Advances in Neural Information Processing Systems}, pages 6840--6851, 2020.

\bibitem[Honkavaara et~al.(2013)Honkavaara, Saari, Kaivosoja, P{\"o}l{\"o}nen, Hakala, Litkey, M{\"a}kynen, and Pesonen]{honkavaara2013processing}
Eija Honkavaara, Heikki Saari, Jere Kaivosoja, Ilkka P{\"o}l{\"o}nen, Teemu Hakala, Paula Litkey, Jussi M{\"a}kynen, and Liisa Pesonen.
\newblock Processing and assessment of spectrometric, stereoscopic imagery collected using a lightweight uav spectral camera for precision agriculture.
\newblock \emph{Remote Sensing}, 5\penalty0 (10):\penalty0 5006--5039, 2013.

\bibitem[Huang et~al.(2021)Huang, Savkin, and Huang]{huang2021decentralized}
Hailong Huang, Andrey~V Savkin, and Chao Huang.
\newblock Decentralized autonomous navigation of a uav network for road traffic monitoring.
\newblock \emph{IEEE Transactions on Aerospace and Electronic Systems}, 57\penalty0 (4):\penalty0 2558--2564, 2021.

\bibitem[Huang et~al.(2022)Huang, Chen, and Huang]{huang2022ufpmp}
Yecheng Huang, Jiaxin Chen, and Di Huang.
\newblock Ufpmp-det: Toward accurate and efficient object detection on drone imagery.
\newblock In \emph{Proceedings of the AAAI conference on Artificial Intelligence}, pages 1026--1033, 2022.

\bibitem[Jahn et~al.(2021)Jahn, Rombach, and Ommer]{jahn2021high}
Manuel Jahn, Robin Rombach, and Bj{\"o}rn Ommer.
\newblock High-resolution complex scene synthesis with transformers.
\newblock In \emph{Proceedings of the IEEE/CVF International Conference on Computer Vision Workshops}, pages 7054--7065, 2021.

\bibitem[Karras et~al.(2019)Karras, Laine, and Aila]{karras2019style}
Tero Karras, Samuli Laine, and Timo Aila.
\newblock A style-based generator architecture for generative adversarial networks.
\newblock In \emph{Proceedings of the IEEE/CVF Conference on Computer Vision and Pattern Recognition}, pages 4401--4410, 2019.

\bibitem[Khanam and Hussain(2024)]{khanam2024yolov11}
Rahima Khanam and Muhammad Hussain.
\newblock Yolov11: An overview of the key architectural enhancements.
\newblock \emph{arXiv preprint arXiv:2410.17725}, 2024.

\bibitem[Khose et~al.(2024)Khose, Pal, Agarwal, Deepanshi, Hoffman, and Chattopadhyay]{khose2024skyscenes}
Sahil Khose, Anisha Pal, Aayushi Agarwal, Deepanshi, Judy Hoffman, and Prithvijit Chattopadhyay.
\newblock Skyscenes: A synthetic dataset for aerial scene understanding.
\newblock In \emph{Proceedings of the European Conference on Computer Vision}, pages 19--35, 2024.

\bibitem[Kingma et~al.(2021)Kingma, Salimans, Poole, and Ho]{kingma2021variational}
Diederik Kingma, Tim Salimans, Ben Poole, and Jonathan Ho.
\newblock Variational diffusion models.
\newblock In \emph{Advances in Neural Information Processing Systems}, pages 21696--21707, 2021.

\bibitem[Kingma and Welling(2013)]{kingma2013auto}
Diederik~P Kingma and Max Welling.
\newblock Auto-encoding variational bayes.
\newblock \emph{arXiv preprint arXiv:1312.6114}, 2013.

\bibitem[Li et~al.(2025{\natexlab{a}})Li, Zhao, Wang, Xu, and Zhu]{li2025remdet}
Chen Li, Rui Zhao, Zeyu Wang, Huiying Xu, and Xinzhong Zhu.
\newblock Remdet: Rethinking efficient model design for uav object detection.
\newblock In \emph{Proceedings of the AAAI Conference on Artificial Intelligence}, pages 4643--4651, 2025{\natexlab{a}}.

\bibitem[Li et~al.(2025{\natexlab{b}})Li, Chen, Liu, Gao, Hong, Yeung, Lu, and Jia]{li2025trackdiffusion}
Pengxiang Li, Kai Chen, Zhili Liu, Ruiyuan Gao, Lanqing Hong, Dit-Yan Yeung, Huchuan Lu, and Xu Jia.
\newblock Trackdiffusion: Tracklet-conditioned video generation via diffusion models.
\newblock In \emph{Proceedings of the IEEE/CVF Winter Conference on Applications of Computer Vision}, pages 3539--3548, 2025{\natexlab{b}}.

\bibitem[Li et~al.(2020)Li, Wang, Wu, Chen, Hu, Li, Tang, and Yang]{li2020generalized}
Xiang Li, Wenhai Wang, Lijun Wu, Shuo Chen, Xiaolin Hu, Jun Li, Jinhui Tang, and Jian Yang.
\newblock Generalized focal loss: Learning qualified and distributed bounding boxes for dense object detection.
\newblock In \emph{Advances in Neural Information Processing Systems}, pages 21002--21012, 2020.

\bibitem[Li et~al.(2023)Li, Liu, Wu, Mu, Yang, Gao, Li, and Lee]{li2023gligen}
Yuheng Li, Haotian Liu, Qingyang Wu, Fangzhou Mu, Jianwei Yang, Jianfeng Gao, Chunyuan Li, and Yong~Jae Lee.
\newblock Gligen: Open-set grounded text-to-image generation.
\newblock In \emph{Proceedings of the IEEE/CVF Conference on Computer Vision and Pattern Recognition}, pages 22511--22521, 2023.

\bibitem[Li et~al.(2021)Li, Wu, Koh, Tang, and Sun]{li2021image}
Zejian Li, Jingyu Wu, Immanuel Koh, Yongchuan Tang, and Lingyun Sun.
\newblock Image synthesis from layout with locality-aware mask adaption.
\newblock In \emph{Proceedings of the IEEE/CVF International Conference on Computer Vision}, pages 13819--13828, 2021.

\bibitem[Lin et~al.(2014)Lin, Maire, Belongie, Hays, Perona, Ramanan, Doll{\'a}r, and Zitnick]{lin2014microsoft}
Tsung-Yi Lin, Michael Maire, Serge Belongie, James Hays, Pietro Perona, Deva Ramanan, Piotr Doll{\'a}r, and C~Lawrence Zitnick.
\newblock Microsoft coco: Common objects in context.
\newblock In \emph{Proceedings of the European Conference on Computer Vision}, pages 740--755, 2014.

\bibitem[Nichol and Dhariwal(2021)]{nichol2021improved}
Alexander~Quinn Nichol and Prafulla Dhariwal.
\newblock Improved denoising diffusion probabilistic models.
\newblock In \emph{Proceedings of the International Conference on Machine Learning}, pages 8162--8171, 2021.

\bibitem[Redmon and Farhadi(2018)]{redmon2018yolov3}
Joseph Redmon and Ali Farhadi.
\newblock Yolov3: An incremental improvement.
\newblock \emph{arXiv preprint arXiv:1804.02767}, 2018.

\bibitem[Redmon et~al.(2016)Redmon, Divvala, Girshick, and Farhadi]{redmon2016you}
Joseph Redmon, Santosh Divvala, Ross Girshick, and Ali Farhadi.
\newblock You only look once: Unified, real-time object detection.
\newblock In \emph{Proceedings of the IEEE/CVF Conference on Computer Vision and Pattern Recognition}, pages 779--788, 2016.

\bibitem[Ren et~al.(2016)Ren, He, Girshick, and Sun]{ren2016faster}
Shaoqing Ren, Kaiming He, Ross Girshick, and Jian Sun.
\newblock Faster r-cnn: Towards real-time object detection with region proposal networks.
\newblock \emph{IEEE Transactions on Pattern Analysis and Machine Intelligence}, 39\penalty0 (6):\penalty0 1137--1149, 2016.

\bibitem[Rezende et~al.(2014)Rezende, Mohamed, and Wierstra]{rezende2014stochastic}
Danilo~Jimenez Rezende, Shakir Mohamed, and Daan Wierstra.
\newblock Stochastic backpropagation and approximate inference in deep generative models.
\newblock In \emph{Proceedings of the International Conference on Machine Learning}, pages 1278--1286, 2014.

\bibitem[Rizzoli et~al.(2023)Rizzoli, Barbato, Caligiuri, and Zanuttigh]{rizzoli2023syndrone}
Giulia Rizzoli, Francesco Barbato, Matteo Caligiuri, and Pietro Zanuttigh.
\newblock Syndrone-multi-modal uav dataset for urban scenarios.
\newblock In \emph{Proceedings of the IEEE/CVF International Conference on Computer Vision}, pages 2210--2220, 2023.

\bibitem[Rombach et~al.(2022)Rombach, Blattmann, Lorenz, Esser, and Ommer]{rombach2022high}
Robin Rombach, Andreas Blattmann, Dominik Lorenz, Patrick Esser, and Bj{\"o}rn Ommer.
\newblock High-resolution image synthesis with latent diffusion models.
\newblock In \emph{Proceedings of the IEEE/CVF Conference on Computer Vision and Pattern Recognition}, pages 10684--10695, 2022.

\bibitem[Rovira-M{\'a}s et~al.(2008)Rovira-M{\'a}s, Zhang, and Reid]{rovira2008stereo}
Francisco Rovira-M{\'a}s, Qin Zhang, and John~F Reid.
\newblock Stereo vision three-dimensional terrain maps for precision agriculture.
\newblock \emph{Computers and Electronics in Agriculture}, 60\penalty0 (2):\penalty0 133--143, 2008.

\bibitem[Shen et~al.(2023)Shen, Lee, Kwon, and Bhattacharyya]{shen2023progressive}
Yi-Ting Shen, Hyungtae Lee, Heesung Kwon, and Shuvra~S Bhattacharyya.
\newblock Progressive transformation learning for leveraging virtual images in training.
\newblock In \emph{Proceedings of the IEEE/CVF Conference on Computer Vision and Pattern Recognition}, pages 835--844, 2023.

\bibitem[Sohl-Dickstein et~al.(2015)Sohl-Dickstein, Weiss, Maheswaranathan, and Ganguli]{sohl2015deep}
Jascha Sohl-Dickstein, Eric Weiss, Niru Maheswaranathan, and Surya Ganguli.
\newblock Deep unsupervised learning using nonequilibrium thermodynamics.
\newblock In \emph{Proceedings of the International Conference on Machine Learning}, pages 2256--2265, 2015.

\bibitem[Song et~al.(2021)Song, Meng, and Ermon]{song2020denoising}
Jiaming Song, Chenlin Meng, and Stefano Ermon.
\newblock Denoising diffusion implicit models.
\newblock In \emph{Proceedings of the International Conference on Learning Representations}, pages 14205--14224, 2021.

\bibitem[Song and Ermon(2020)]{song2020improved}
Yang Song and Stefano Ermon.
\newblock Improved techniques for training score-based generative models.
\newblock In \emph{Advances in Neural Information Processing Systems}, pages 12438--12448, 2020.

\bibitem[Tang et~al.(2025)Tang, Cao, Wu, Li, Yao, Bai, Jiang, Li, and Meng]{tang2025aerogen}
Datao Tang, Xiangyong Cao, Xuan Wu, Jialin Li, Jing Yao, Xueru Bai, Dongsheng Jiang, Yin Li, and Deyu Meng.
\newblock Aerogen: Enhancing remote sensing object detection with diffusion-driven data generation.
\newblock In \emph{Proceedings of the IEEE/CVF Conference on Computer Vision and Pattern Recognition}, pages 3614--3624, 2025.

\bibitem[Tian et~al.(2025)Tian, Ye, and Doermann]{tian2025yolov12}
Yunjie Tian, Qixiang Ye, and David Doermann.
\newblock Yolov12: Attention-centric real-time object detectors.
\newblock \emph{arXiv preprint arXiv:2502.12524}, 2025.

\bibitem[Toker et~al.(2024)Toker, Eisenberger, Cremers, and Leal-Taix{\'e}]{toker2024satsynth}
Aysim Toker, Marvin Eisenberger, Daniel Cremers, and Laura Leal-Taix{\'e}.
\newblock Satsynth: Augmenting image-mask pairs through diffusion models for aerial semantic segmentation.
\newblock In \emph{Proceedings of the IEEE/CVF Conference on Computer Vision and Pattern Recognition}, pages 27695--27705, 2024.

\bibitem[Vahdat and Kautz(2020)]{vahdat2020nvae}
Arash Vahdat and Jan Kautz.
\newblock Nvae: A deep hierarchical variational autoencoder.
\newblock In \emph{Advances in Neural Information Processing Systems}, pages 19667--19679, 2020.

\bibitem[Wang et~al.(2024{\natexlab{a}})Wang, Darrell, Rambhatla, Girdhar, and Misra]{wang2024instancediffusion}
Xudong Wang, Trevor Darrell, Sai~Saketh Rambhatla, Rohit Girdhar, and Ishan Misra.
\newblock Instancediffusion: Instance-level control for image generation.
\newblock In \emph{Proceedings of the IEEE/CVF Conference on Computer Vision and Pattern Recognition}, pages 6232--6242, 2024{\natexlab{a}}.

\bibitem[Wang et~al.(2024{\natexlab{b}})Wang, Gao, Chen, Zhou, Cai, Hong, Li, Jiang, Yeung, Xu, et~al.]{wang2024detdiffusion}
Yibo Wang, Ruiyuan Gao, Kai Chen, Kaiqiang Zhou, Yingjie Cai, Lanqing Hong, Zhenguo Li, Lihui Jiang, Dit-Yan Yeung, Qiang Xu, et~al.
\newblock Detdiffusion: Synergizing generative and perceptive models for enhanced data generation and perception.
\newblock In \emph{Proceedings of the IEEE/CVF Conference on Computer Vision and Pattern Recognition}, pages 7246--7255, 2024{\natexlab{b}}.

\bibitem[Wu et~al.(2024)Wu, Chen, and Wang]{wu2024domain}
Ke Wu, Jiaxin Chen, and Miao Wang.
\newblock Domain adaptive object detection for uav-based images by robust representation learning and multiple pseudo-label aggregation.
\newblock In \emph{Proceedings of the ACM MM Workshops on Efficient Multimedia Computing under Limited}, page 59–67, 2024.

\bibitem[Wu et~al.(2023)Wu, Zhao, Chen, Gu, Zhao, He, Zhou, Shou, and Shen]{wu2023datasetdm}
Weijia Wu, Yuzhong Zhao, Hao Chen, Yuchao Gu, Rui Zhao, Yefei He, Hong Zhou, Mike~Zheng Shou, and Chunhua Shen.
\newblock Datasetdm: Synthesizing data with perception annotations using diffusion models.
\newblock In \emph{Advances in Neural Information Processing Systems}, pages 54683--54695, 2023.

\bibitem[Xie et~al.(2023)Xie, Li, Huang, Liu, Zhang, Zheng, and Shou]{xie2023boxdiff}
Jinheng Xie, Yuexiang Li, Yawen Huang, Haozhe Liu, Wentian Zhang, Yefeng Zheng, and Mike~Zheng Shou.
\newblock Boxdiff: Text-to-image synthesis with training-free box-constrained diffusion.
\newblock In \emph{Proceedings of the IEEE/CVF International Conference on Computer Vision}, pages 7452--7461, 2023.

\bibitem[Yang et~al.(2023)Yang, Wang, Gan, Li, Lin, Wu, Duan, Liu, Liu, Zeng, et~al.]{yang2023reco}
Zhengyuan Yang, Jianfeng Wang, Zhe Gan, Linjie Li, Kevin Lin, Chenfei Wu, Nan Duan, Zicheng Liu, Ce Liu, Michael Zeng, et~al.
\newblock Reco: Region-controlled text-to-image generation.
\newblock In \emph{Proceedings of the IEEE/CVF Conference on Computer Vision and Pattern Recognition}, pages 14246--14255, 2023.

\bibitem[Yim et~al.(2024)Yim, Lee, Eum, Shen, Zhang, Kwon, and Bhattacharyya]{yim2024synplay}
Jinsub Yim, Hyungtae Lee, Sungmin Eum, Yi-Ting Shen, Yan Zhang, Heesung Kwon, and Shuvra~S Bhattacharyya.
\newblock Synplay: Importing real-world diversity for a synthetic human dataset.
\newblock \emph{arXiv e-prints}, pages arXiv--2408, 2024.

\bibitem[Zhang et~al.(2023)Zhang, Rao, and Agrawala]{zhang2023adding}
Lvmin Zhang, Anyi Rao, and Maneesh Agrawala.
\newblock Adding conditional control to text-to-image diffusion models.
\newblock In \emph{Proceedings of the IEEE/CVF International Conference on Computer Vision}, pages 3836--3847, 2023.

\bibitem[Zhang et~al.(2021)Zhang, Ling, Gao, Yin, Lafleche, Barriuso, Torralba, and Fidler]{zhang2021datasetgan}
Yuxuan Zhang, Huan Ling, Jun Gao, Kangxue Yin, Jean-Francois Lafleche, Adela Barriuso, Antonio Torralba, and Sanja Fidler.
\newblock Datasetgan: Efficient labeled data factory with minimal human effort.
\newblock In \emph{Proceedings of the IEEE/CVF Conference on Computer Vision and Pattern Recognition}, pages 10145--10155, 2021.

\bibitem[Zhao et~al.(2023)Zhao, Sheng, Bao, Chen, Chen, Wen, Yuan, Liu, Zhou, Chu, et~al.]{zhao2023x}
Hanqing Zhao, Dianmo Sheng, Jianmin Bao, Dongdong Chen, Dong Chen, Fang Wen, Lu Yuan, Ce Liu, Wenbo Zhou, Qi Chu, et~al.
\newblock X-paste: Revisiting scalable copy-paste for instance segmentation using clip and stablediffusion.
\newblock In \emph{Proceedings of the International Conference on Machine Learning}, pages 42098--42109, 2023.

\bibitem[Zheng et~al.(2023)Zheng, Zhou, Li, Qi, Shan, and Li]{zheng2023layoutdiffusion}
Guangcong Zheng, Xianpan Zhou, Xuewei Li, Zhongang Qi, Ying Shan, and Xi Li.
\newblock Layoutdiffusion: Controllable diffusion model for layout-to-image generation.
\newblock In \emph{Proceedings of the IEEE/CVF Conference on Computer Vision and Pattern Recognition}, pages 22490--22499, 2023.

\bibitem[Zhou et~al.(2024)Zhou, Li, Ma, Zhang, and Yang]{zhou2024migc}
Dewei Zhou, You Li, Fan Ma, Xiaoting Zhang, and Yi Yang.
\newblock Migc: Multi-instance generation controller for text-to-image synthesis.
\newblock In \emph{Proceedings of the IEEE/CVF Conference on Computer Vision and Pattern Recognition}, pages 6818--6828, 2024.

\bibitem[Zhu et~al.(2024)Zhu, Li, Liu, Yuan, Huang, Shan, and Ma]{zhu2024odgen}
Jingyuan Zhu, Shiyu Li, Yuxuan~Andy Liu, Jian Yuan, Ping Huang, Jiulong Shan, and Huimin Ma.
\newblock Odgen: Domain-specific object detection data generation with diffusion models.
\newblock In \emph{Advances in Neural Information Processing Systems}, pages 63599--63633, 2024.

\end{thebibliography}
}

\clearpage
\setcounter{page}{1}
\setcounter{section}{0}
\renewcommand\thesection{\Alph{section}}
\setcounter{figure}{0}
\setcounter{table}{0}
\renewcommand{\thefigure}{\Alph{figure}}
\renewcommand{\thetable}{\Alph{table}}

\maketitlesupplementary

In this document, we additionally provide an implementation details in~\cref{sec:supple_implementation}, supplementary ablation study on our method in~\cref{sec:supple_ablation}, comparison of inference time in~\cref{sec:supple_inference} and detailed algorithm of our proposed label refinement method in~\cref{sec:algo_label}.

\section{More Implementation Details}
\label{sec:supple_implementation}
\subsection{Implementation of UAVGen model}
During training process of UAVGen, we initialized the object layout fuse module in Visual Prototype layout encoder and MLP, Gate Attention module in text layout encoder with Kaiming Uniform, while remaining parameters of UAVGen were initialized with FLUX.1-dev, a pre-trained diffusion model. While the remained parameters were frozen, we finetuned the parameters of ControlNet and layout encoder (both Visual Prototype layout encoder and the text layout encoder) with Adam optimizer and constant learning rate of 1e-5 for 30k steps. The batch size is set to 8 and images are resized to 512$\times$512. $w$ is set to 2, when we employ the foreground-aware reweighted loss.

\subsection{Dataset Generation Statistics}
On the VisDrone~\cite{du2019visdrone} benchmark, we generated 18,649 image patches with VPC-DM, which were merged into 738 full-resolution images. On the UAVDT~\cite{du2018unmanned} benchmark, our method generated 29,403 image patches with VPC-DM, which were merged into 9,802 full-resolution images. For all compared methods, we generated 6,474 and 24,143 images for VisDrone and UAVDT respectively, matching the size of each original training set.

\subsection{Training process of Object Detector}
We choose GFL \cite{li2020generalized} as general object detection model and RemDet \cite{li2025remdet} as UAV-based object detection model to evaluate the effect of synthetic data. GFL was trained for 12 epochs via the SGD optimizer which is configured with a momentum of 0.9 and a weight decay factor of 0.0001. We set the batch size to 16 and set the initial learning rate to 0.005, with a linear warm-up and decreased by 10 times at 8th and 11th epoch. During inference, background bounding boxes were filtered out using a confidence threshold of 0.05, and NMS was applied with an IoU threshold of 0.5, limiting the maximum number of bounding boxes to 3000. The input image sizes were set to 1333$\times$800 on VisDrone \cite{du2019visdrone} and UAVDT \cite{du2018unmanned}. All experiments were conducted on 8 NVIDIA RTX 3080Ti GPUs.

In terms of UAV-based object detection model, following the default experimental set of Remdet \cite{li2025remdet}, we trained the detector on VisDrone dataset for 300 epochs, with a learning rate of 0.01, and applied data augmentation techniques such as mixup and Mosaic. The input image sizes were set to 640$\times$640 according to default. All experiments were conducted on 8 NVIDIA RTX 3080Ti GPUs.

\section{More Ablation Study}
\label{sec:supple_ablation}
\subsection{Ablation on Initial Detector}
To validate the initial detector’s robustness under data scarcity, we conducted an ablation study on initial detector Faster R-CNN~\cite{ren2016faster} with 10\%, 50\%, and 100\% training data (Table~\ref{tab:ablation_scale}). Note that mAP* denotes the initial detector’s performance, while mAP, AP$_{50}$, and AP$_{75}$ reflect enhancements for the GFL~\cite{li2020generalized} detector. With 10\% training data, the initial detector’s mAP* drops from 21.5\% to 13.7\%, while mAP stays at 25.6\% (0.3\% lower than the 100\% baseline). Notably, 50\% data yields the highest AP$_{50}$ (45.1\%), slightly outperforming the baseline. These results confirm our method’s strong performance and robustness under data scarcity.

\begin{table}[h]
    \centering
    \caption{Ablation study on initial detector for full pipeline. ``mAP*'' shows initial detector mAP.}
    \begin{tabular}{c|c|ccc}
        \toprule
        Scale & mAP* & mAP & AP$_{50}$ & AP$_{75}$ \\
        \midrule
        10\%  & 13.7 & 25.6 & 44.4 & 25.7 \\
        50\%  & 19.8 & 25.9 & \textbf{45.1} & 25.9 \\
        \underline{100\%} & \underline{21.5} & \underline{\textbf{25.9}} & \underline{44.8} & \underline{\textbf{26.0}} \\
        \bottomrule
    \end{tabular}
    \label{tab:ablation_scale}
\end{table}

To verify the effectiveness of the Label Refinement (LR) module when paired with weak initial detectors, we conducted ablation experiments using different detector architectures, including the lightweight weak YOLOv3-tiny~\cite{redmon2018yolov3}, as well as the commonly used Faster R-CNN~\cite{ren2016faster}, which serves as our default detector. As specified in the experiment, mAP* specifically represents the mAP performance of the initial detector, while other metrics (mAP, AP$_{50}$, AP$_{75}$) reflect argument for enhancing the GFL~\cite{li2020generalized} detector. 
As shown in Table~\ref{tab:ablation_label-refinement}, for the weaker YOLOv3-tiny, its mAP* is significantly lower (10.1\%) compared to the baseline Faster R-CNN (21.5\%). Nonetheless, LR still delivers consistent accuracy improvements, with mAP, AP$_{50}$, and AP$_{75}$ all increasing by 0.2\% respectively, confirming LR’s effectiveness even for detectors with notably reduced initial performance. Collectively, our method works effectively with detectors of different accuracies, demonstrating its effectiveness and practical application value.

\begin{table}[h]
    \centering
    \caption{Ablation study on initial detector for Label Refinement. ``mAP*'' shows initial detector mAP.}
    \begin{tabular}{c|c|ccc}
        \toprule
        Detector & mAP* & mAP & AP$_{50}$ & AP$_{75}$ \\
        \midrule
        w/o LR & - & 25.5 & 44.5 & 25.6 \\
        YOLOv3-tiny & 10.1 & 25.7 & 44.7 & 25.8 \\
        \underline{Faster R-CNN} & \underline{21.5} & \underline{\textbf{25.9}} & \underline{\textbf{44.8}} & \underline{\textbf{26.0}} \\
        \bottomrule
    \end{tabular}
    \label{tab:ablation_label-refinement}
\end{table}

\subsection{Ablation on Text Prompt}
Both prompts are detailed in Section 3.2 (Multi-Source Condition Encoding): the Global Prompt is fed into the diffusion model as a scene-level condition, while each Individual Prompt is embedded and fused with its corresponding spatial position to form object-level layout conditions. These layout conditions are then aggregated into Layout Embedding, which is injected into ControlNet to guide generation.
To clarify their respective contributions, we conducted an ablation study, as show in Table~\ref{tab:ablation_text-prompt}, which shows that: (1) using only the global prompt yields coarse scene coherence but poor object fidelity; (2) using only individual prompts improves object details but lacks global consistency; and (3) combining both achieves the best balance, validating our design.

\begin{table}[h]
    \caption{Ablation study on text prompt}
    \centering
    \begin{tabular}{cc ccc}
        \toprule
        $t^g_i$ & $t^{c_j}$ & mAP & AP$_{50}$ & AP$_{75}$ \\
    \midrule
        \Checkmark &  & 24.1 & 42.4 & 24.0 \\
         & \Checkmark & 25.1 & 43.6 & \textbf{25.2} \\
        \Checkmark & \Checkmark & \textbf{25.2} & \textbf{43.8} & 25.1 \\
        \bottomrule
    \end{tabular}
    \label{tab:ablation_text-prompt}
\end{table}

\subsection{Ablation on Cross-dataset Enhancement}
We further validated the cross-dataset generalization and practicality of our method through a cross-dataset augmentation experiment. Specifically, we trained our UAVGen on the UAVDT~\cite{du2018unmanned} dataset, and use the synthetic data it generates to augment the training process of GFL~\cite{li2020generalized} detector on the VisDrone~\cite{du2019visdrone} dataset. As presented in Table~\ref{tab:ablation_cross-enhancement}, compared with training only on real VisDrone data, introducing cross-dataset generated samples brings clear and stable gains: mAP improves by 0.3\% from 24.5\% to 24.8\%, and AP$_{50}$ rises by 1.2\% from 42.1\% to 43.3\%. Such improvements confirm that our method can effectively boost detection performance under low-data or domain-shift conditions, showing strong generalization ability across different UAV datasets.

\begin{table}[h]
    \caption{Ablation study on cross-dataset enhancement.}
    \centering
    \begin{tabular}{l ccc}
        \toprule
            Method & mAP & AP$_{50}$ & AP$_{75}$ \\
        \midrule
            Real Only (VisDrone) & 24.5 & 42.1 & 24.5 \\       
            w/ UAVGen (UAVDT) & \textbf{24.8} & \textbf{43.3} & \textbf{24.8} \\
        \bottomrule
    \end{tabular}
    \label{tab:ablation_cross-enhancement}
\end{table}

\section{Inference time}
\label{sec:supple_inference}
We evaluated inference time on a single NVIDIA H20 GPU. UAVGen takes 8.8 seconds per image for the synthesis process, while AeroGen takes 3.1 seconds per image. We regard such moderate computational overhead as acceptable, especially considering the significant performance improvements brought to detector training. For instance, UAVGen outperforms AeroGen by 1.0\% on the VisDrone benchmark and 1.6\% on the UAVDT benchmark, while reducing the total detector training time by approximately 2 hours.

\section{Algorithm of Label Refinement}
\label{sec:algo_label}
Algorithm \ref{alg:label_refinement} presents the detailed implementation of the Label Refinement in Focal Region-Enhanced Data Pipeline introduced in the main text, aiming to address label inaccuracies (missing annotations, false generations, and label misalignments) in the dense synthetic dataset \(\mathcal{D}^{\text{syn\_dense}}\). Taking \(\mathcal{D}^{\text{syn\_dense}}\), a pre-trained detector \(D(\cdot)\), task-specific thresholds (\(\tau^{\text{ref}}, \alpha, \beta, \gamma\)) as inputs, the algorithm first obtains detector-derived pseudo-labels \(L^{\text{det}}\) for each synthetic image, then performs IoU-based matching between original labels \(L^{\text{real}}\) and \(L^{\text{det}}\) to build matched pairs \(\mathcal{M}\). Subsequently, it handles missing generations via relaxed IoU threshold and confidence filtering (as in \cref{eq:refine1}), supplements high-quality false generations from \(L^{\text{det}}\) (as in \cref{eq:refine2}), and corrects label misalignments using high-confidence detector boxes (as in \cref{eq:refine3}), ultimately outputting the refined dataset \(\mathcal{D}^{\text{ref}}\) with optimized labels.

\begin{algorithm*}[t]
    \caption{Label Refinement.}
    \label{alg:label_refinement}
    \setcounter{AlgoLine}{0}
    \LinesNumbered
    \SetKw{KwDict}{Dictionary}
    \KwIn{Synthesized dataset $\mathcal{D}^{\text{syn\_dense}}$, pre-trianed detector $D(\cdot)$, thresholds $\tau^{\text{ref}},~\alpha,~\beta,~\gamma$
    }
    \KwOut{Synthesized dataset $\mathcal{D}^{\text{ref}}$ with refined labels}

    $\mathcal{D}^{\textnormal{ref}}\gets \{\}$~~
    \textcolor{gray}{\texttt{\# Initialize the refined dataset}}

    \ForEach{$(I^{\textnormal{syn}}, L^{\textnormal{real}})\in \mathcal{D}^{\textnormal{syn\_dense}}$}{
        \textcolor{gray}{\texttt{\# Obtain detected boxes via a pre-trained detector as~\cref{eq:label_detect}}}\\
        $L^{\textnormal{det}} \gets D(I^{\textnormal{syn}})$~~
        \textcolor{gray}{\texttt{\# $L^{\textnormal{det}}$ contains bounding box $b_j^{\textnormal{det}}$, class $c_j$, and confidence $s_j$}}

        \BlankLine

        \textcolor{gray}{\texttt{\# Perform IoU-based matching with~\cref{eq:pairs}}}\\
        \KwDict $\mathcal{M} \gets \{\}$~~
        \textcolor{gray}{\texttt{\# Maintain $\mathcal{M}$ as a dictionary for the convenience of usage}}
        
        \ForEach{$(b_i^{\textnormal{real}}, c_i^{\textnormal{real}}) \in L^{\textnormal{real}}$}{
            \ForEach{$(b_j^{\textnormal{det}}, c_j^{\textnormal{det}}, s_j) \in L^{\textnormal{det}}$}{
            \If{$\textnormal{IoU}(b_i^{\textnormal{real}}, b_j^{\textnormal{det}})\ge \tau^{\textnormal{ref}}$}{
                $\mathcal{M}[b_i^{\textnormal{real}}]\gets b_j^{\textnormal{det}}$~~
                \textcolor{gray}{\texttt{\# Storage the matched pairs $(b_i^{\textnormal{real}},~b_j^{\textnormal{det}})$}}
            }
            }
        }

        \BlankLine

        $L^{\textnormal{ref}} \gets \{\}$~~
        \textcolor{gray}{\texttt{\# Start to refine the labels}}

        \BlankLine

        \textcolor{gray}{\texttt{\# Process missed generations as~\cref{eq:refine1}}}\\

        \ForEach{$(b_i^{\textnormal{real}},~c_i^{\textnormal{real}}) \in L^{\textnormal{real}}$}{
            \If{$b_i^{\textnormal{real}}\in \mathcal{M}\textnormal{.keys()}$~~\textnormal{and}~~$s_j \ge \Phi^{-1}_{c_j}(\alpha)$ }{
                    \textcolor{gray}{\texttt{\# Retain the matched labels with high confidence scores}}\\
                $L^{\textnormal{ref}}\gets L^{\textnormal{ref}} \cup (b_i^{\textnormal{real}}, ~c_i^{\textnormal{real}})$\\
            }     
        }

        \BlankLine

        \textcolor{gray}{\texttt{\# Process false generations as~\cref{eq:refine2}}}\\
        
        \ForEach{$(b_j^{\textnormal{det}},~c_j^{\textnormal{det}},~s_j) \in L^{\textnormal{det}}$}{
            \If{$b_j^{\textnormal{det}}\notin \mathcal{M}\textnormal{.values()}~~\textnormal{and}~~s_j\ge \Phi^{-1}_{c_j}(\beta)$}{
                \textcolor{gray}{\texttt{\# Include the well generated regions to the synthetic dataset}}\\
                    $L^{\textnormal{ref}} \gets L^{\textnormal{ref}} \cup \{(b_j^{\textnormal{det}}, c_j)\}$
            }
        }

        \BlankLine

        \textcolor{gray}{\texttt{\# Process label misalignments as~\cref{eq:refine3}}}\\
        \ForEach{$(b_i^{\textnormal{ref}},~c_i^{\textnormal{ref}}) \in L^{\textnormal{ref}}$}{
            $b_j^{\textnormal{det}}\gets \mathcal{M}[b_i^{\textnormal{ref}}]$~~
            \textcolor{gray}{\texttt{\# Get the paired detected region}}\\
            \If{$s_j>\Phi^{-1}_{c_j}(\gamma)$}{
                \textcolor{gray}{\texttt{\# Refine the label with high-confidence detections}}\\
                $L^{\textnormal{ref}} \gets L^{\textnormal{ref}} / \{(b_i^{\textnormal{ref}}, c_i^{\textnormal{ref}})\}$,~~
                $L^{\textnormal{ref}} \gets L^{\textnormal{ref}} \cup \{(b_j^{\textnormal{det}}, c_j^{\textnormal{det}})\}$\\
            }
        }

        $\mathcal{D}^{\textnormal{ref}}\gets \mathcal{D}^{\textnormal{ref}}\cup (I^{\textnormal{syn}}, L^{\textnormal{ref}})$
    }
    
    \Return $\mathcal{D}^{\text{ref}}$
\end{algorithm*}

\end{document}